\documentclass[sigconf]{acmart}
\usepackage[utf8]{inputenc}
\usepackage{amsmath, enumitem, graphicx, hyperref, lipsum, multirow, array}

\title{Unsupervised Cycle Detection in Agentic Applications}

\author{Felix George}
\affiliation{
  \institution{IBM Research}
  \country{India}
}
\email{felix.george@ibm.com}

\author{Harshit Kumar}
\affiliation{
  \institution{IBM Research}
  \country{India}
}
\email{harshit.kumar@ibm.com}

\author{Divya Pathak}
\affiliation{
  \institution{IBM Research}
  \country{India}
}
\email{divya.pathak@ibm.com}

\author{Kaustabha Ray}
\affiliation{
  \institution{IBM Research}
  \country{India}
}
\email{kaustabha.ray@ibm.com}

\author{Mudit Verma}
\affiliation{
  \institution{IBM Research}
  \country{India}
}
\email{mudit.verma@ibm.com}

\author{Pratibha Moogi}
\affiliation{
  \institution{IBM Research}
  \country{India}
}
\email{pratibha.moogi@ibm.com}

\date{September 2025}
\begin{document}

\begin{abstract}
Agentic applications powered by Large Language Models exhibit non-deterministic behaviors that can form hidden execution cycles, silently consuming resources without triggering explicit errors. Traditional observability platforms fail to detect these costly inefficiencies. We present an unsupervised cycle detection framework that combines structural and semantic analysis. Our approach first applies computationally efficient temporal call stack analysis to identify explicit loops and then leverages semantic similarity analysis to uncover subtle cycles characterized by redundant content generation. Evaluated on 1575 trajectories from a LangGraph-based stock market application, our hybrid approach achieves an F1 score of 0.72 (precision: 0.62, recall: 0.86), significantly outperforming individual structural (F1: 0.08) and semantic methods (F1: 0.28). While these results are encouraging, there remains substantial scope for improvement, and future work is needed to refine the approach and address its current limitations. 
\end{abstract}

\maketitle
\section{Introduction}
\label{sec:introduction}
Agentic applications powered by Large Language Models (LLMs) enable seamless autonomous operations but introduce critical reliability challenges due to their nondeterministic nature and complex interactions \cite{chan2023harms, hammond2025multiagent, packer2024visibility}. Observability platforms like OpenLLMetry~\cite{OpenLLMetry} collect rich trajectories governing agentic workflows but lack automated detection of inefficiencies in generated trajectories. One of the fundamental inefficiencies with agentic trajectories is hidden cycles, repetitive action sequences that consume resources without progress or explicit errors, quite different from explicit cycles that involves repeated operation calls. Both inflate operational costs through excessive token consumption, as LLM inference on high-end GPUs is far more expensive than microservice resources like CPU or memory. A key challenge with agentic systems is thus differentiating between \textbf{productive cycles} and \textbf{bad cycles}. A productive cycle occur when an agent or tool is repeatedly invoked but  makes sequential progress at each step towards its goal. In contrast, bad cycles arise when an agent executes tools or sub-agents redundantly, without yielding  additional insights or progress.

Trajectory analysis in agentic systems is increasingly gaining interest in the research community. For instance, the Multi Agent System Failure Taxonomy(MAST) in "Why Do Multi-Agent LLM Systems Fail?"~\cite{cemri2025multi} empirically identifies 14 Multi-Agent Systems failure modes, including "Step Repetition" as a symptom of cycles.
Similarly, "SentinelAgent"~\cite{he2025sentinelagentgraphbasedanomalydetection} models interactions as dynamic graphs for LLM-powered oversight, focusing on security anomalies like prompt injections rather than cost-inefficient cycles, and uses supervised matching of known issues. Existing observability platforms (e.g., Datadog~\cite{DatadogLLMObs}, Langfuse~\cite{Langfuse}) monitor metrics like latency and token usage with basic outlier detection but overlook structural repetition and semantic redundancy. These approaches, however, lack unsupervised, real-time methods to detect cost-inefficient cycles in production agentic systems. 

Our work addresses this gap by proposing an unsupervised cycle detection framework that analyzes trajectories to identify both explicit and hidden cycles via hybrid structural and semantic analysis. Our key contributions include: 
\begin{itemize}[leftmargin=1em]
    \item A novel unsupervised framework combining structural and semantic analysis for cycle detection in agentic system.
    \item First solution addressing "bad cycles", a critical cost driver in LLM-based applications
    \item Open-source dataset of 1575 labeled agentic trajectories \footnote{This dataset will be released after paper acceptance with organization policies}
    \item Laying foundations for proactive, adaptable observability in evolving agentic applications
\end{itemize}

\noindent
The rest of this paper is organized as follows:
Section \ref{sec:problem_formulation} formally defines the problem and explains the methodology in detail. Section \ref{sec:experiments_and_discussions} outlines the experimental setup and results obtained. Finally, Section \ref{sec:contributions} concludes the work and outlines future work.

\section{Problem Formulation and Approach}
\label{sec:problem_formulation}
This section introduces some notations and provides a formal definition of the bad cycle detection task in an agent execution trajectory, followed by a detailed description of the proposed approach.

An \emph{agent execution trajectory} $\mathcal{T} = \{s_1, s_2, \dots, s_n\}$ is a collection of spans, where each span $s_i$  is represented as a tuple, $\langle$ \texttt{trace\_id}, \texttt{span\_id}, \texttt{parent\_span\_id}, \texttt{op}, \texttt{input}, \texttt{output} $\rangle$. Here, $\texttt{trace\_id} \in \mathbb{T}$ uniquely identifies the trajectory, $\texttt{span\_id} \in \mathbb{I}$ uniquely identifies the span, and $\texttt{parent\_span\_id} \in \mathbb{I} \cup \{\varnothing\}$ specifies its parent (or $\varnothing$ if root). The fields $\texttt{input}, \texttt{output} \in \mathbb{D}$ capture data flow, where $\mathbb{I}$ denotes the set of all span identifiers and $\mathbb{D}$ denotes the domain of structured input/output data values.
The detection of bad cycles in $\mathcal{T}$ is formulated as a binary classification problem, defined as follows:
$f( \mathcal{T} ) \rightarrow \{0, 1\}$,
where:
\begin{itemize}
    \item \( f( \mathcal{T} ) = 1 \), indicates that \( \mathcal{T} \) contains a bad cycle.
    \item \( f( \mathcal{T} ) = 0 \), indicates that \( \mathcal{T} \) represents a non-cyclical execution or a productive cycle. 
\end{itemize}
We leverage a Directed Acyclic Graph and Call Stack to represent a trajectory $\mathcal{T}$ defined below:
\begin{figure}
    \centering
    \includegraphics[width=0.6\linewidth, angle=90]{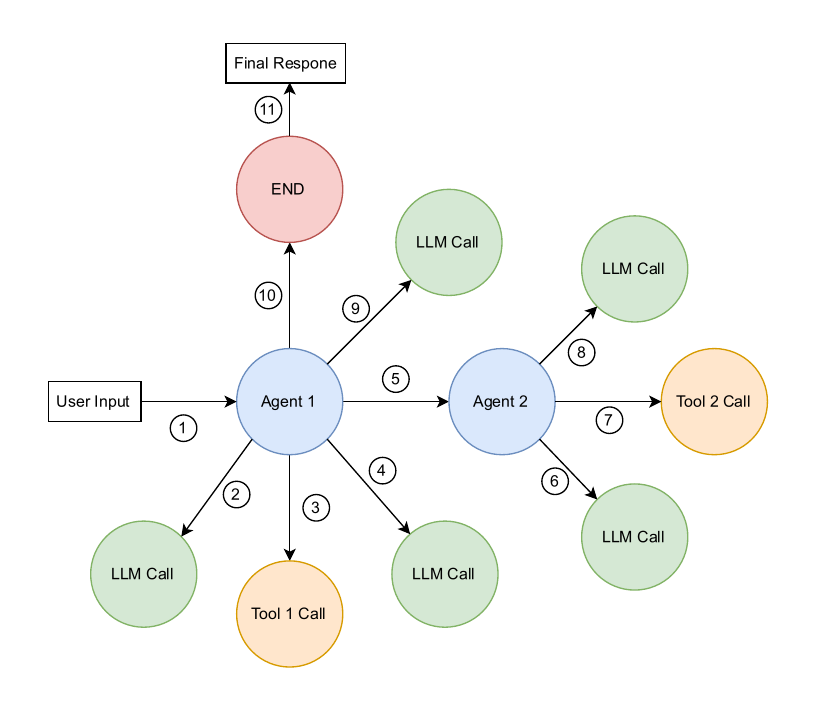}
    \caption{Agent execution trajectory  represented as span tree where nodes  represent spans and edges represent the parent-child relationship. Edge label represent the temporal order.}
    \label{fig:agent_trajectory}
\end{figure}
\begin{figure}
    \centering
    \includegraphics[width=0.7\linewidth]{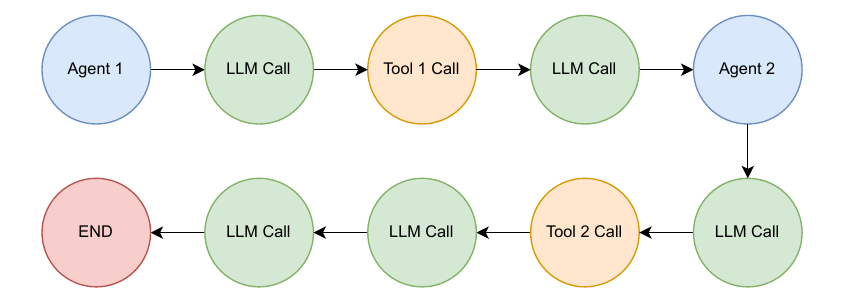}
    \caption{Call Stack representation of Figure~\ref{fig:agent_trajectory}}
    \label{fig:call_stack}
\end{figure}
\begin{figure}
    \centering
    \includegraphics[width=0.5\linewidth]{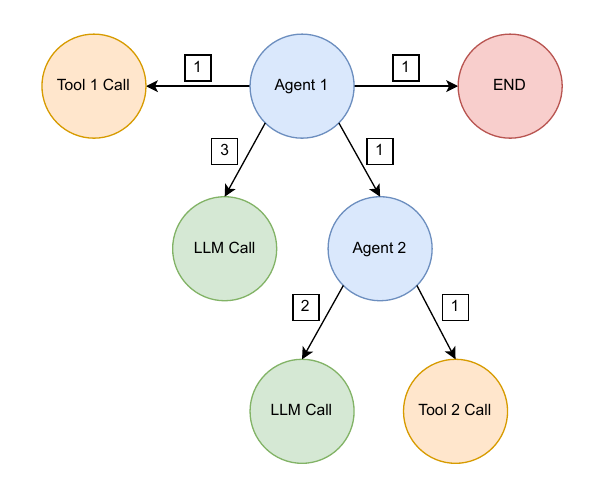}
    \caption{DAG representation of Figure~\ref{fig:agent_trajectory}}
    \label{fig:dag}
\end{figure}
\begin{itemize}[leftmargin=1em]
    \item \textbf{Directed Acyclic Graph (DAG)} 
    For a particular trajectory $\mathcal{T}$,
    we define a DAG $G_\mathcal{T} = (\mathcal{S}, E, w)$, where $\mathcal{S} = \mathcal{T}$ is the set of spans, 
    $E = \{(s_i, s_j) \mid \texttt{parent\_span\_id}(s_j) = \texttt{span\_id}(s_i)\}$ represents parent-child edges, 
    and $w: E \to \mathbb{N}$ assigns a weight corresponding to the frequency of occurrence of each $(s_i, s_j)$ tuple during execution capturing the hierarchical relationships between spans for efficient traversal as shown in Figure \ref{fig:dag}. 

    \item \textbf{Call Stack Representation}
    Let $\mathcal{T}$ be ordered by creation time $t(s_i)$, yielding a sequence 
    $\mathcal{C}_\mathcal{T} = \langle s_{\pi(1)}, s_{\pi(2)}, \dots, s_{\pi(n)} \rangle$ with $t(s_{\pi(1)}) \le t(s_{\pi(2)}) \le \dots \le t(s_{\pi(n)})$. 
    This linear sequence captures the temporal progression of the agent's execution and exposes recurring patterns via subsequences as shown in Figure \ref{fig:call_stack}. 
\end{itemize}

\noindent
In the subsequent section, we  discuss our methodology in detail.

\subsection{Detailed Methodology}
\label{sec:detailed_methodology}

\noindent
Our methodology revolves around three main themes: structural overlap, identifying recurring span subsequences; semantic similarity, measuring output similarity of sibling spans to flag redundancy and a hybrid method that combine both structural and semantic cues to capture cycles that may not be evident when considering either dimension in isolation. 

\noindent
\paragraph{\textbf{Cycle Detection Using DAG representation (CDDAG):}} 
\label{sec:cddag}
Recall that, $G_\mathcal{T} = (\mathcal{S}, E, w)$ is the DAG representation of a trajectory $\mathcal{T}$, where $w(e)$ denotes the frequency of an edge $e \in E$. Define the multi-set of edge weights as $F = \{ w(e) \mid e \in E \}$. The mean $\mu$ and standard deviation $\sigma$ of the edge weights are:
\[
\mu = \frac{1}{|E|} \sum_{e \in E} w(e), \quad
\sigma = \sqrt{\frac{1}{|E|} \sum_{e \in E} (w(e) - \mu)^2}.
\]

An edge $e \in E$ is classified as \emph{cyclic} if its weight exceeds a threshold determined by a tunable parameter $m > 0$:
\[
w(e) > \mu + m \cdot \sigma.
\]

The trajectory $\mathcal{T}$ is then labeled by the classification function
\[
f(\mathcal{T}) =
\begin{cases}
1 & \text{if } \exists e \in E \text{ such that } w(e) > \mu + m \cdot \sigma,\\
0 & \text{otherwise},
\end{cases}
\]
where $f(\mathcal{T}) = 1$ indicates the presence of a bad cycle, and $f(\mathcal{T}) = 0$ indicates a non-cyclical or healthy trajectory.

\paragraph{\textbf{Cycle Detection Using Call Stack representation (CDCS):}} 
\label{sec:cdcs}
Let the trajectory $\mathcal{T}$ be represented as a call stack $\mathcal{C}_\mathcal{T} = [s_1, s_2, \dots, s_n]$, where each $s_i$ is a span. 
A \emph{repeating sequence} is a contiguous subsequence $S = [s_i, s_{i+1}, \dots, s_{i+m-1}]$ of length $m > 2$ that appears multiple times in $\mathcal{C}_\mathcal{T}$. 
Let $w(S)$ denote the frequency of sequence $S$ in $\mathcal{C}_\mathcal{T}$. Define the multi-set of all 
frequencies as $F = \{ w(S) \mid S \in \mathcal{S} \}$ (using a sliding window of variable size to get subsequences and update frequency of subsequence to a Map data structure),
where $\mathcal{S}$ is the set of all contiguous subsequences of $\mathcal{C}_\mathcal{T}$. The mean $\mu$ and standard deviation $\sigma$ of the subsequence frequencies are:
\[
\mu = \frac{1}{|\mathcal{S}|} \sum_{S \in \mathcal{S}} w(S), \quad
\sigma = \sqrt{\frac{1}{|\mathcal{S}|} \sum_{S \in \mathcal{S}} (w(S) - \mu)^2}.
\]

\noindent
A subsequence $S \in \mathcal{S}$ is classified as \emph{cyclic} if its frequency exceeds a threshold determined by a tunable parameter $k > 0$:
\[
w(S) > \mu + k \cdot \sigma.
\]

The trajectory $\mathcal{T}$ is then labeled by
\[
f(\mathcal{T}) =
\begin{cases}
1 & \text{if } \exists S \in \mathcal{S} \text{ such that } w(S) > \mu + k \cdot \sigma, \\
0 & \text{otherwise},
\end{cases}
\]
where $f(\mathcal{T}) = 1$ indicates the presence of a bad cycle, and $f(\mathcal{T}) = 0$ indicates a non-cyclical or productive trajectory. 

Both methods described above aim to identify cycles within $\mathcal{T}$. However, when a tool or agent is invoked repeatedly with different arguments, these methods are rendered unsuitable since they are context unaware and focus solely on the structure of the trajectory. To avert misclassifying such executions, we utilize the output context information recorded in the spans.
\paragraph{\textbf{Cycle detection using semantic analysis (CDSA):}}
\label{sec:cdsa} 
Cyclic spans show high \texttt{input–output} similarity. We use cosine similarity between span outputs to assess trajectory utility, however, this is an expensive computation with $\binom{n}{2}$ comparisons.
We restrict similarity checks to sibling nodes in the DAG of $\mathcal{T}$, reducing cost to $\binom{\log(n)}{2}$, since information flows upward from leaves to parents.
A subgraph is flagged as a bad cycle if a node’s content exceeds similarity threshold $s$ with its sibling since highly similar leaf nodes likely yield similar parents and ancestors.
Let spans of $\mathcal{T}$ be $ S = \{s_1, s_2, \dots, s_n\} $, where the output of each span $ s_i $ is represented as a vector $ \mathbf{v}_i \in \mathbb{R}^d $ in a embedding space of $d$ (any embedding models, e.g. OpenAITextEmbedding~\cite{OpenAIEmbedding}, Qwen Embedding~\cite{Qwen3Embedding} can be leveraged for this purpose). The cosine similarity between two spans $ s_i $ and $ s_j $ quantifies the repetition of information and is defined as: $$\text{cos}(\mathbf{v}_i, \mathbf{v}_j) = \frac{\mathbf{v}_i \cdot \mathbf{v}_j}{\|\mathbf{v}_i\| \|\mathbf{v}_j\|}$$
We define a subgraph of the DAG as exhibiting a bad cycle if there exists a node $ v_i $ with a sibling node $ v_j $ (i.e., both share the same parent) such that their cosine similarity exceeds a predefined threshold $ \phi \in (0, 1] :\text{cos}(\mathbf{v}_i, \mathbf{v}_j) > \phi $
and $\mathcal{T}$ is labeled as:
\[
    f:( \mathcal{T} ) = 
        \begin{cases} 
        1 & \text{cos}(\mathbf{v}_i, \mathbf{v}_j) > \phi, \\
        0 & \text{otherwise}.
        \end{cases}
\]
\paragraph{\textbf{Hybrid Approach}:}
We utilize the above in conjunction in a hybrid multi-stage approach to label a trajectory as follows:
\begin{itemize}[leftmargin=1em]
    \item \textbf{Call Stack Analysis}: Examines the sequence of function calls within the agent's execution to identify potential cyclic patterns, leveraging call stack structure.
    \item \textbf{Semantic Similarity Confirmation}: Upon detection of a potential cycle through call stack analysis, the semantic similarity between trajectory spans is computed to confirm the presence of repetitive content.
\end{itemize}
This hybrid approach, which combines call stack-based cycle detection with semantic similarity analysis, provides a computationally efficient unsupervised method to detect cycles in agentic trajectories.
\section{Experiments and Discussion}
\label{sec:experiments_and_discussions}
In this section, we describe our experimental setup, dataset and label generation and the results obtained.
\begin{table}[t]
\caption{Dataset Specifications}
\label{tab:dataset_specs}
\scalebox{0.85}{
\begin{tabular}{|l|c|}
\hline
\textbf{Attribute} & \textbf{Value} \\
\hline
Application Framework & LangGraph \\
\hline
User Prompts & 525 \\
\hline
System Prompt Types & 3 (Poor, Good, Strict) \\
\hline
Total Agentic Trajectories & 1575 \\
\hline
Prompt Classes & 6 \\
\hline
Number of Classes & 2 \\
\hline
Bad cycles & 57 \\
\hline
\end{tabular}
}
\end{table}
\subsection{Dataset and Ground Truth Creation}

We built a Stock market agentic AI application using langgraph with a hierarchical agent architecture consisting of a supervisor, search, and stock agent, with internet search and Yahoo Finance APIs as agent tools. The goal of the agentic AI application is to generate agentic trajectories and establish a ground truth dataset for benchmarking models. We created 525 unique user prompts clustered into 6 prompt classes (share price, comparison, analysis, forecast, news, trends) to ensure trajectory diversity. Combined with three system prompt configurations - poor, good (inspired by ReAct\cite{yao2023react} prompting ), and strict - it resulted in 1575 agent trajectories. The three system prompt configurations were carefully chosen to emulate the user prompt-engineering strategies. 

\begin{figure}[ht]
    \centering
    \includegraphics[width=0.9\linewidth]{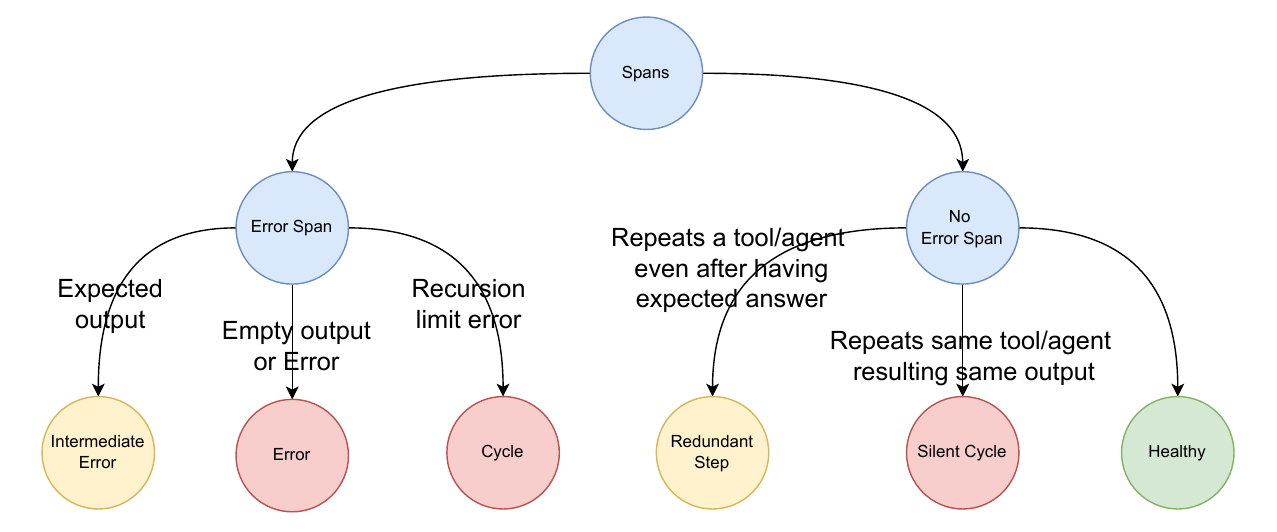}
    \caption{Ground truth creation strategy for trajectory dataset}
    \label{fig:ground_truth}
\end{figure}
Figure~\ref{fig:ground_truth} illustrates the process to label the trajectories, which identifies spans by errors, output quality, and repetition, starting with error detection. If an error span is present and is attributed to a recursion limit, the trajectory is classified as an \textit{Error Cycle}. For error trajectories without output or empty output, the label is \textit{Error}. However, if the agent recovers from the error and ultimately produces the correct output, it is labeled as \textit{Intermediate Error}. If no errors are found, we manually detect redundant steps in trajectories as follows: trajectories with unnecessary steps despite available data are labeled \textit{Redundant Step}. Similarly, if the agent repeatedly performs an action that regenerates information it already has, leading to identical outputs, it is labeled as \textit{Silent Cycle}. Finally, trajectories where the agent generates the expected answer efficiently are deemed \textit{Productive}. Trajectories belonging to either the Error Cycle or the Silent Cycle category are classified as bad cycles, the rest of the cycles are labeled as good/healthy. 
This structured approach ensures consistent and reliable labeling, facilitating robust analysis of agent performance. 
\begin{table*}[ht]
\caption{Classification Results for Cycle Detection Methods with the best parameter values}
\label{tab:results}
\scalebox{0.9}{
\begin{tabular}{|l|c|c|ccc|ccc|}
\hline
\multirow{2}{*}{\textbf{Method}} & \multirow{2}{*}{\textbf{Threshold}} & \multirow{2}{*}{\textbf{Accuracy}} & \multicolumn{3}{c|}{\textbf{Cycle}} & \multicolumn{3}{c|}{\textbf{Non Cycle}} \\
\cline{4-9}
& & & \textbf{Precision} & \textbf{Recall} & \textbf{F1-score} & \textbf{Precision} & \textbf{Recall} & \textbf{F1-score} \\
\hline
CDDAG & $\mu + 1.4 \cdot \sigma$ & 0.65 & 0.05 & 0.44 & 0.08 & 0.97 & 0.65 & 0.78 \\
\hline
CDCS & $\mu + 0.5 \cdot \sigma$ & 0.92 & 0.30 & 0.88 & 0.45 & 1.00 & 0.92 & 0.96 \\
\hline
CDSA & $s > 0.85$ & 0.83 & 0.16 & 0.91 & 0.28 & 1.00 & 0.82 & 0.90 \\
\hline
Hybrid Approach & $s > 0.83, \mu + 0.5 \cdot \sigma$ & 0.98 & 0.62 & 0.86 & 0.72 & 0.99 & 0.98 & 0.99 \\
\hline
\end{tabular}
}
\end{table*}
\subsection{Results and Discussion}
\begin{figure*}
    \centering
    \includegraphics[scale = 0.4]{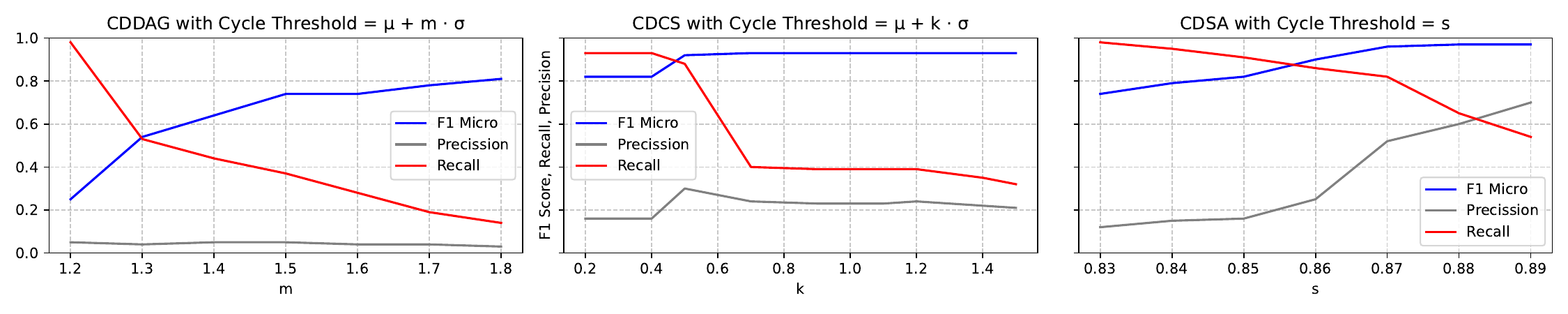}
    \caption{F1-Micro, Precesion and recall for structural approaches CDDAG, CDCS, and semantic similarity approach CDSA}
    \label{fig:combined_results}
\end{figure*}

This susbsection compares the four approaches, CDDAG, CDCS, CDSA and Hybrid approach, on the labeled trajectory dataset prepared in the previous sub section. Figure~\ref{fig:combined_results} shows the performance of each of the methods CDDAG, CDCS and CDSA when the parameters m, k, and s are varied. Table~\ref{tab:results} shows best results achieved using all four methods.

As shown in the classification report, refer Table~\ref{tab:results}, CDDAG method performed poorly, achieving an F1 score of only $0.08$ (for $k=1.4$ for the "Cycle" class due to the low precision; refer to the leftmost illustration in Figure~\ref{fig:combined_results}. In further investigation, we observed that the CDDAG method has a large number of false positives ($525$) and false negatives ($32$), resulting in a very low precision of 0.05 and a recall of 0.44 with $m=1.4$ for $m$ in $\mu + m \cdot \sigma$. For $m=1.5$, we get an increase in false negatives of ($36$) but it decreases the number of false positives to ($367$). As shown in Figure~\ref{fig:combined_results}, the threshold parameter $m$ was varied in  between $1.2$ and $1.8$, in steps of $0.1$. At $m=1.2$, it had one false negative but a large number of false positives (1179), while at $m=1.8$ it yielded the highest false negatives (49) but the fewest false positives (240). On further investigation of misclassified agent trajectories, we found that \textbf{frequent edge traversals alone are not reliable indicators of cycles, as common non-cyclic agentic patterns can also appear frequently.}\\

Using only the CDCS approach, the best F1-score for the task of cycle detection was $0.45$ at $k=0.5$, with high recall (0.88) but low precision (0.30). We varied $k$ between $0.2$ and $1.5$  wherein Figure~\ref{fig:combined_results} depicts the trade-off between false negatives (as low as $7$) and false positives (up to $114$). Although this method is effective at capturing temporal pattern repetition, it is susceptible to misclassify non-cyclic behaviors. \textbf{Purely structural representation is thus not sufficient for accurate cycle identification.}

Furthermore, the CDSA method, which relies solely on semantic analysis (Figure~\ref{fig:combined_results}), achieves an F1-score of 0.28 with very high recall ($0.91$), correctly identifying 52 of 57 bad cycles. However, its low precision ($0.16$) and the large number of false positives ($267$ at $s>0.85$) indicate that, while semantic similarity captures most cycles, it frequently misclassifies productive cycles as bad. This limitation arises primarily from the large time-series data generated by the \textit{timeseries\_daily} tool of the \textit{stock agent}, where high cosine similarity values occur even when the underlying time series of two companies differ substantially. \textbf{As a result, the current semantic similarity approach—well-suited for string comparisons—fails to effectively compare numerical time-series data, leading to trajectory misclassifications}. Addressing this challenge will require the development of novel semantic similarity techniques that can handle both string and numerical time-series data, which is an interesting future improvement that should yield better results.
The low precision of the structural approaches (CDDAG and CDCS) and semantic method (CDSA), despite their high recall, \textbf{highlights the complexity of distinguishing between good and healthy trajectories from cyclic agentic behavior}. This underscores the need for a hybrid approach that can take advantage of the strengths of each method while mitigating their weaknesses.

Results in Table~\ref{tab:results} shows that the proposed hybrid methodology significantly improves overall performance. It achieves an F1-score of 0.72 for the "Cycle" class and 0.99 for "Non Cycle" class, with a precision of 0.62 and a recall of 0.86. This result demonstrates a marked improvement over the individual methods, particularly in reducing false positives, thereby making the framework practical for real-world applications. It also provides a significant computational advantage by reducing the number of trajectories that require the expensive semantic similarity calculation. This hybrid method  captures both explicit and silent cycles, offering a robust and efficient solution for a critical problem in agentic application development.

\section{Conclusions and Future Work}
\label{sec:contributions}

In this paper we  present a novel unsupervised cycle detection framework for agentic applications. Our work addresses a critical, emerging challenge posed by the non-deterministic and autonomous nature of these systems, specifically due to  elusive hidden unproductive cycles. We demonstrate that traditional structural analysis alone, whether via call stack or DAG representations, is insufficient to reliably detect all forms of cyclical behavior, particularly the hidden cycles.

Looking ahead, our vision is to evolve this framework into a comprehensive anomaly detection system, adept at identifying a wide array of issues, including data drift, latent errors, and beyond, while expanding our data set to encompass a richer diversity of agent behaviors, large language models, and user prompts to enhance the generalizability of our insights. We plan to integrate real-time cycle and anomaly detection capabilities into agentic applications, alongside advanced semantic comparison techniques surpassing the limitations of cosine similarity for large timeseries JSON data to refine our methodology. Additionally, during the creation of ground truth, we have incorporated additional trajectory labels, allowing the reuse of this data set as we expand our approach to address a broader spectrum of anomalies. As part of this ground truth creation, we leveraged a classification methodology as outlined earlier. As part of a robust evaluation strategy, we plan to compare multiple approaches to the ground truth labeling via machine learning approaches and then compare our approach with these multi-faceted ground truth sets.

\bibliographystyle{ACM-Reference-Format}
\bibliography{main}
\end{document}